\documentclass{article}

\usepackage{microtype}
\usepackage{subfigmat}

\usepackage{hyperref}

\usepackage[accepted]{icml2019}

\usepackage{times}
\usepackage{soul}
\usepackage{url}
\usepackage[utf8]{inputenc}
\usepackage[small]{caption}
\usepackage{graphicx}
\usepackage{booktabs}
\usepackage{algorithm}
\usepackage{algorithmic}
\usepackage{here}
\usepackage{natbib}
\usepackage{amssymb,amsmath,amsthm,amsfonts}
\usepackage{color}
\usepackage{tikz}
\usepackage{url}
\usepackage{mathtools}
\usepackage{here}
\usepackage{dcolumn}
\usepackage{booktabs}
\usetikzlibrary{arrows, decorations.markings,shapes.geometric, positioning}

\date{}
\newtheorem{theo}{Theorem}

  \newtheorem{coro}{Corollary}

\newcommand{\Eqref}[1]{Equation~(\ref{#1})}
\newcommand{\figref}[1]{figure~\ref{#1}}
\newcommand{\Figref}[1]{Figure~\ref{#1}}
\newcommand{\theoref}[1]{Theorem~\ref{#1}}

\newcommand{\Cororef}[1]{Corollary~\ref{#1}}

\newcommand{\E}{\mathbb{E}}
\tikzstyle{myvec}=[line width=1mm,draw=gray,-triangle 45,postaction={draw, line width=3mm, shorten >=4mm, -}]

\newcommand{\memo}[1]{}
\newcommand{\comm}[1]{}

\icmltitlerunning{Optimistic Proximal Policy Optimization}

\begin{document}
\twocolumn[
\icmltitle{Optimistic Proximal Policy Optimization}



\icmlsetsymbol{equal}{*}
\begin{icmlauthorlist}
\icmlauthor{Takahisa Imagawa}{aist}
\icmlauthor{Takuya Hiraoka}{aist,nec}
\icmlauthor{Yoshimasa Tsuruoka}{aist,ut}
\end{icmlauthorlist}

\icmlaffiliation{aist}{National Institute of Advanced Industrial Science and Technology, Tokyo, Japan}
\icmlaffiliation{nec}{NEC Central Research Laboratories, Kanagawa, Japan}
\icmlaffiliation{ut}{The University of Tokyo, Tokyo, Japan}
\icmlcorrespondingauthor{Takahisa Imagawa}{imagawa.t@aist.go.jp}

\icmlkeywords{Reinforcement Learning, Exploration, Uncertainty}

\vskip 0.3in
]



\printAffiliationsAndNotice{}  

\begin{abstract}
Reinforcement Learning, a machine learning framework for training an autonomous agent based on rewards, has shown outstanding results in various domains.
However, it is known that learning a good policy is difficult in a domain where rewards are rare.
We propose a method, optimistic proximal policy optimization (OPPO) to alleviate this difficulty.
OPPO considers the uncertainty of the estimated total return and optimistically evaluates the policy based on that amount.
We show that OPPO outperforms the existing methods in a tabular task.\memo{これで良い？}

\end{abstract}

\section{Introduction}
Reinforcement learning is a framework to learn a good policy in terms of total expected extrinsic rewards by interacting with an environment.
It has shown super-human performance in the game of Go and in Atari games~\cite{mnih2015human, silver2017mastering}.
In the early days, RL algorithms such as Q-learning, and state-action-reward-state-action (SARSA)~\cite{sutton1998introduction}, and recently, more sophisticated algorithms have been proposed.
Among the latter, proximal policy optimization (PPO) is one of the most popular algorithms, because it can be used in a variety of tasks such as Atari games and robotic control tasks~\cite{schulman2017proximal}.

However, learning a good policy is difficult when the agent rarely receives extrinsic rewards.  
Existing methods alleviate this problem by adding another type of reward called intrinsic reward.
For example, as an intrinsic reward, \citet{pathak2017curiosity} and \citet{burda2019large} use prediction error of the next state, and \citet{burda2019exploration} use evaluation of state novelty.
However, these methods are not based on solid theoretical backgrounds.

Uncertainty Bellman exploration (UBE) is another method to alleviate the sparse reward problem, which has a more solid theoretical background~\cite{o2018uncertainty}. 
UBE evaluates the value of a policy higher when the estimation of the value is more uncertain, 
like in ``optimism in face of uncertainty'' in multi-armed bandit problems~\cite{bubeck2012regret}.
\citet{o2018uncertainty} showed a relationship between the local uncertainty and the uncertainty of the expected return and applied the uncertainty estimation to SARSA.
%
\memo{uncertaintyにaはつけない方が良いのでは？UBEと英文校正確認}

We apply the idea of UBE to PPO and propose a new algorithm named optimistic PPO (OPPO) which evaluates the uncertainty of the total return of a policy and updates the policy in the same way as PPO.
By updating the policy like PPO, its policy is expected to be stable, and this allows OPPO to evaluate the uncertainty of estimated values in states that are far from the current state.

\section{Background}\label{sec:notations}
\subsection{Uncertainty Bellman Equation and Exploration}
Markov decision processes (MDPs) are models of sequential decision-making problems.
In this paper, we focus on an MDP with a finite horizon, state, and action space.
An MDP is defined as a tuple, $\langle \mathcal{S}, \mathcal{A}, r, T, \rho, H \rangle$, where $\mathcal{S}$ is a set of possible states, $\mathcal{A}$ is a set of possible actions;
and $r$ is a reward function $\mathcal{S} \times \mathcal{A} \rightarrow \mathbb{R}$, which defines the expected reward when the action is taken at the state; 
$T$ is a transition function $\mathcal{S} \times \mathcal{A} \times \mathcal{S} \rightarrow [0,1]$, which defines the transition probability to the next state when the action is taken at the current state; 
$\rho$ is a probability distribution of the initial state, and
$H \in \mathbb{N}$ is the horizon length of the MDP, i.e. the number of actions until the end of an episode. 

The objective of an agent/learner is to learn a good policy in terms of expected total return.
Formally, policy $\pi_\theta(a|s)$ $(s \in \mathcal{S}, a \in \mathcal{A})$ is the probability of taking action $a$ at state $s$, where $\theta$ is a set of parameters that determines the probability (for the sake of simplicity, we often omit $\theta$).
The Q-value $Q^{h, \pi}_{(s,a)}$, $(Q^{H+1, \pi}_{(s,a)} := 0)$ is an expected total return when the agent is at state $s$, time-step $h$, takes action $a$, and follows policy $\pi$ after taking action $a$.

Let us assume the Bayesian setting of Q-value estimation, where there are priors and posteriors over the mean reward function $r$ and the transition function $T$.
Let $\hat{r}$ be the sampled reward function, $\hat{T}$ be the sampled transition function from prior or posterior, and $\mathcal{F}_\tau$ be the sigma-algebra of all data (e.g. states, actions, rewards) earned by $\tau$ times sampling.
It is known that there exists a unique $\hat{Q}^{h, \pi}_{(s,a)}$ that satisfies the Bellman equation,
\begin{equation}
\hat{Q}^{h, \pi}_{(s,a)} =\hat{r}(s,a)+\sum_{s^{\prime}, a^{\prime}} \pi(a|s) \hat{T}(s,a,s^{\prime}) \hat{Q}^{h+1, \pi}_{(s',a')},
\end{equation}
for all $s$ and $a$, for $h = 0, \dots, H$, where $\hat{Q}^{H+1,\pi}_{(s,a)} = 0$.
\citet{o2018uncertainty} extend this Bellman equation to the variance/uncertainty of $\hat{Q}^{h, \pi}_{(s,a)}$.

To prove theoretical results,
let us assume that the state transition of the MDP is a directed acyclic graph (DAG) and that expected reward $r(s,a)$ is bounded for all states and actions.
We denote the conditional variance of a random variable $x$ as
\begin{equation}
\mathbf{var}_\tau x := \E ((x - \E(x|\mathcal{F}_\tau)|\mathcal{F}_\tau)^2.
\end{equation}
We denote the maximum of Q-value as $Q_{\max}$ and $\nu_{\tau}(s,a)$ as
\begin{align}
\mathbf{v a r}_{\tau} \hat{r}(s,a) + {Q_{\max}}^{2} \sum_{s^{\prime}} \frac{\mathbf{v a r}_{\tau} \hat{T}(s,a,s') } {{T}_{\tau}(s, a, s')},
\end{align}
where ${T}_{\tau}(s, a, s') := \E_{\hat{T}} [ \hat{T}(s,a,s') | \mathcal{F}_{\tau}]$.
The Q-value satisfies the following equation~\cite{o2018uncertainty}.
\begin{theo}
For any policy $\pi$, there exists a unique $Q^{h,\pi}_{2,\tau}$ that satisfies the uncertainty Bellman equation,

\begin{equation}
\small{Q^{h,\pi}_{2,\tau}(s,a)=\nu_{\tau}(s, a)+\sum_{s', a'} \pi(a'|s') {T}_{\tau}(s, a, s') Q_{2,\tau}^{h+1,\pi}(s^{\prime}, a^{\prime})}\label{eq:uncertaintyBellman}
\end{equation}

for all $(s, a)$ and $h = 1, \dots, H$, where $Q^{H+1,\pi}_{2,\tau} = 0$, and $Q_{2,\tau}^{h, \pi} \geq  \mathbf{v a r}_{\tau} \hat{Q}^{h,\pi}$ point-wise.
\label{theo:uncertaintyBellman}
\end{theo}
This theorem shows a relationship between the local uncertainty, $\nu_{\tau}(s, a)$ and the uncertainty of estimated Q-values.

For convenience of discussion in later sections, we introduce some notations.
Let us denote the solution of the Bellman equation, 
\begin{equation}
\small{Q^{h,\pi}_{1,\tau}(s,a)=r_{\tau}(s, a)+\sum_{s', a'} \pi(a'|s') {T}_{\tau}(s, a, s') Q_{1,\tau}^{h+1,\pi}(s^{\prime}, a^{\prime})}\label{eq:Bellman}
\end{equation}
as $Q^{h,\pi}_{1,\tau}$, where the estimated mean reward, ${r}_{\tau}(s, a)$ is $\E_{\hat{r}} [ \hat{r}(s,a) | \mathcal{F}_{\tau}]$.
For $i = 1, 2$,
\begin{align}
V^{h,\pi}_{i, \tau} (s) &:= \sum_a \pi(a|s) Q_{i, \tau}^{h, \pi}(s, a), \\
A^{h,\pi}_{i, \tau}(s,a) &:= Q_{i, \tau}^{h,\pi}(s, a) - V_{i, \tau}^{h,\pi} (s),\\
\eta_{i, \tau}(\pi) &:= \sum_s \rho(s) V_{i,\tau}^{0,\pi}(s).
\end{align}

To estimate $\nu_\tau(s,a)$, \citet{o2018uncertainty} start from the case where the domain is tabular. 
Let $n_{s,a}$ denote the number of times action $a$ is chosen at state $s$ and let $\sigma_{r}^{2}$ denote the variance of a reward sampled from the reward distribution.
We assume that the reward distribution and its prior is Gaussian, and the prior over the transition function is Dirichlet; then 
\begin{align}
\mathbf{ v a r }_{\tau} \hat{r}(s,a) &\leq \sigma_{r}^{2} / n_{s a},\\
\sum_{s'}\mathbf{ v a r }_{\tau} \hat{T}(s,a,s')/ T_\tau(s,a,s') &\leq |\mathcal{S}_{s,a}| / n_{s a},
\end{align}
where $|\mathcal{S}_{s,a}|$ is the number of next states reachable from $(s, a)$.
\comm{（AAAIまでに）限定的な問題設定と思われない様に何か，もう少し正当性を言いたい，CLTとか，どういう意味で妥当と言えば良いのか，木佐森さん山崎さんあたりに聞いてみる，}
\memo{UBEの議論に寄せる　．．を仮定する．これは少なくとも．．．で妥当}
Thus, there exists a constant $C_u$ which satisfies $\nu_\tau(s,a) \leq \frac{C_u}{n_{s,a}}$, e.g. $C_u = \sigma_{r}^2 + Q_{\max}^2 |\mathcal{S}_{s,a}|$.
Since this exact upper bound is too loose in most cases, UBE heuristically chooses $C_u$ instead of using the parameter assured to satisfy the bound. 
In a domain other than the tabular, UBE extends the discussion above and uses pseudo-counts to estimate the local uncertainty.
\citet{o2018uncertainty} applied UBE to SARSA~\cite{sutton1998introduction}, which is a more primitive algorithm than Proximal Policy Optimization.

\subsection{Proximal Policy Optimization}\label{sec:ppo}
Proximal Policy Optimization (PPO) is a simplified version of trust region policy optimization (TRPO)\footnote{While the original TRPO and PPO are formulated under the assumption that the policy is run for an MDP with an infinite horizon, they have recently been extended in the case of finite horizon~\cite{azizzadenesheli2018trust}, which is the same setting as ours.}. 
Although TRPO shows promising results in control tasks~\cite{schulman2015trust}, PPO empirically shows better results in most cases~\cite{schulman2017proximal}.
PPO uses a clipped variable as follows, so as not to change policy drastically.
\begin{equation}
L(\theta)=\bar{\mathbb{E}}_{h}\left[\min \left(l_{h}(\theta) \bar{A}^{h}, \operatorname{clip}\left(l_{h}(\theta), 1-\epsilon, 1+\epsilon\right) \bar{A}^{h}\right)\right],
\label{def:clip}
\end{equation}
where $\theta$ is the parameters of the policy,
$h$ is time-step,
$l_h(\theta)$ is
$\frac{\pi_{\theta}\left(a_{h} | s_{h}\right)}{\pi_{\theta_{\text { old }}}\left(a_{h} | s_{h}\right)}$,
$\bar{A}^{h}$ is the estimated advantage value, e.g. the estimated value of $A^{h,\pi}_{1, \tau}(s_{h},a_{h})$ in this paper,
and $\bar{\mathbb{E}}_{h}[\cdot]$ is the empirical average over a batch of samples.
The clipping function $\operatorname{clip}\left(x, 1-\epsilon, 1+\epsilon\right)$ means $x = 1+\epsilon $ if $x > 1+\epsilon$ and $x = 1-\epsilon $ if $x < 1-\epsilon$. 
PPO samples the data by executing actions for $T$ time-steps following the policy and repeating it $N$ times.
PPO updates the policy by maximizing $[L - \text{prediction error of V-value} + \text{entropy of policy}]$ in the data. 

\subsection{Exploration Based on Intrinsic Reward}
Random network distillation (RND) is recently proposed for alleviating the problem of sparse reward~\cite{burda2019exploration}.
It has shown outstanding performance in Atari games.
RND uses two neural networks called a target network $f_t$ and a predictor network $f_p$.
Each network maps state/observation $x$ to its value $f_t(x)$ or $f_p(x)$.
The networks are randomly initialized, and the target network's parameters are fixed, on the other hand, the predictor learns the outputs of the target.
The intrinsic reward for observation $x$ is defined as the difference of output $||f_t(x) - f_p(x)||^2$.
As a reward, RND uses $[\text{extrinsic one} + \text{intrinsic one}]$, instead of using only the extrinsic one. 
RND uses the reward defined above and learns a policy like PPO.
RND updates the policy to maximize $[\text{PPO's objective} -\text{differences of outputs of the networks}]$ in the batch data.
It is expected that more observations lead to smaller differences of the outputs, which means the intrinsic reward is smaller.
In RND, the intrinsic rewards can be seen as a kind of pseudo-count bonus.
However, there is no theoretical discussion about how this bonus should be used.

There are other methods for exploration by the intrinsic rewards.
To calculate the intrinsic rewards, \citet{bellemare2016unifying} used context tree switching, and \citet{ostrovski2017count} used pixcelCNN.
However, those methods depend on visual heuristics and are not straightforward to apply to other tasks than Atari games, e.g. control tasks whose inputs are sensor data.
\citet{ecoffet2019go} proposed an another method for exploration, which is based on memorization and random search rather than intrinsic reward.
Although it shows state-of-the-art performance on Montezuma's Revenge, it is also not straightforward to extend the method to other tasks.
\citet{tang2017exploration} proposed a method similar to RND which evaluates the state novelty by using a hash function.\memo{RNDの方が良いと言いたいがフェアな比較ではないので保留}

\section{Optimistic Proximal Policy Optimization}\label{sec:oppo}
We propose optimistic proximal policy optimization (OPPO), which is a variant of PPO.
OPPO optimizes a policy based on optimistic evaluation of the expected return where the evaluation is optimistic by the amount of the uncertainty of the expected return. 

First, we explain its theoretical background.
We denote the optimistic value of policy $\tilde{\eta}(\pi)_\tau$ as below:
\begin{align}
\tilde{\eta}_{\tau}(\pi) := \eta_{1, \tau}(\pi) + 2\beta \sqrt{\eta_{2, \tau}(\pi)}, 
\label{def:exact_eta}
\end{align}
where $\beta > 0$ is a hyper-parameter for exploration.
Setting the high value to $\beta$ means emphasizing exploration more than exploitation.
Let us denote the value of policy $\hat{\eta}(\pi)$ as $\sum_{s,a} \rho(s) \pi(a|s) \hat{Q}^{0,\pi}_{(s,a)}$.
Then the following corollary is derived from \theoref{theo:uncertaintyBellman}.
\begin{coro}
\begin{equation}
\mathbf{ v a r }_{\tau} \left(\hat{\eta}(\pi) \right) \leq \eta_{2,\tau}(\pi)
\label{ieq:relation_var-eta2}
\end{equation}
\label{coro:relation_var-eta2}
\end{coro}
This corollary shows that $\eta_{2,\tau}(\pi)$ is an upper bound of the uncertainty of the expected return of $\pi$.
In general, more data lead to more accurate estimation, and this means lower $\nu_\tau(s,a)$ and ${\eta}_{\tau,2}(\pi)$.
Especially if $\nu_\tau(s,a) = 0$, ${\eta}_{\tau,2}(\pi) = 0$.
Also, $0 \leq \mathbf{ v a r }_{\tau} \left(\hat{\eta}(\pi) \right)\leq \eta_{2,\tau}(\pi)$.
Therefore, the difference of  $\mathbf{ v a r }_{\tau} \left(\hat{\eta}(\pi) \right)$ and $\eta_{2,\tau}(\pi)$ decreases to zero as the number of data increases.
These facts show that evaluating $\mathbf{ v a r }_{\tau} \left(\hat{\eta}(\pi) \right)$ by $\eta_{2,\tau}(\pi)$ is reasonable.
Besides, $\eta_{1, \tau}(\pi)$ is an estimation of the mean of $\hat{\eta}(\pi)$.
Thus, $\tilde{\eta}(\pi)_\tau$ is a form that the estimated return plus its uncertainty and seeking a policy which maximizes $\tilde{\eta}(\pi)_\tau$ is reasonable in terms of ``optimism in face of uncertainty''.

However, it is difficult to find policy $\pi'$ which maximizes $\tilde{\eta}_{\tau}(\pi')$ by directly evaluating $\tilde{\eta}_{\tau}(\pi')$.
Thus, following PPO, 
OPPO approximates $\tilde{\eta}_{\tau}(\pi')$ based on the current policy $\pi$.
Let $\mathcal{L}_{\tau}(\pi, \pi')$ denote
\begin{align}
 \tilde{\eta}_{\tau}(\pi) + \sum_{h,s, a}\rho_h^{\pi}(s) \pi'(a|s) \left(  {A}_{1, \tau}^{h,\pi}(s,a) + \beta \frac{{A}_{2, \tau}^{h,\pi}(s,a)}{\sqrt{\eta_{2,\tau}(\pi)}} \right).
\label{def:approx_eta}
\end{align}
Then the following equations are satisfied.
\begin{theo}
For any parameters of policy $\phi$,
\begin{align}
\mathcal{L}_{\tau}(\pi_\phi,\pi_\phi) &=  \tilde{\eta}_{\tau}(\pi_\phi) \label{eq:approx_zero}\\
\nabla_\theta \mathcal{L}_{\tau}(\pi_\phi, \pi_\theta)|_{\theta = \phi} &= \nabla_\theta \tilde{\eta}_{\tau}(\pi_\theta) |_{\theta = \phi} \label{eq:approx_one}
\end{align}
\label{theo:similarity}
\end{theo}
\theoref{theo:similarity} means that $\tilde{\eta}_\tau(\pi')$ can be approximated by $\mathcal{L}_{\tau}(\pi, \pi')$ with enough accuracy if $\pi$ and $\pi'$ are not very different.
\comm{（AAAIまで）TRPOの定理１みたいなことを示せるはずなので，示しておく．}
Therefore, OPPO chooses the next policy $\pi'$ so as to increase the estimated value of $\mathcal{L}_{\tau}(\pi, \pi')$ with regularizing the `similarity' between $\pi$ and $\pi'$ by the clipping function introduced in section~\ref{sec:ppo}.

The objective function of OPPO is the same as $L$ in \Eqref{def:clip}, except that OPPO uses $\tilde{A}^h$ instead of $\bar{A}^h$ in the equation, where $\tilde{A}^h$ is 
\begin{equation}
{A}_1(s_h, a_h) + \beta {A}_2(s_h, a_h) / \sqrt{\eta_2 +c}.
\label{def:oppoobj}
\end{equation}
Parameter $c \geq 0$ is introduced for stabilizing the estimation when $\eta_2(\pi)$ is nearly zero.
Note that \theoref{theo:similarity} is valid if the square root in equations~\eqref{def:exact_eta}, \eqref{def:approx_eta} are either $\sqrt{\eta_{2, \tau}(\pi)}$ or $\sqrt{\eta_{2, \tau}(\pi)+c}$.
The terms, $\eta_2$, ${A}_1(s, a)$ and ${A}_2(s, a)$, are the estimated values of $\eta_{2,\tau}(\pi)$, $A^{h,\pi}_{1, \tau}(s,a)$ and $A^{h,\pi}_{2, \tau}(s,a)$, respectively, which are calculated based on generalized advantage function estimation~\cite{schulman2015high}.
We show the details in ~\ref{app:alg}.
The other parts of the objective function of OPPO are prediction error of V-values and entropy of policy, which are the same as PPO.

Note that simply adding the bonuses ${n_{s,a}}^{-1/2}$ to the extrinsic rewards instead of adding bonuses like UBE and OPPO may be overly optimistic, as shown in an example in \citet{o2018uncertainty},
although ordinary count-based exploration is based on the bonuses~\cite{bellemare2016unifying,ostrovski2017count,tang2017exploration}.

OPPO can be combined with an arbitrary estimator of the local uncertainty.
For example, the local uncertainty can be directly evaluated by bootstrap sampling of the reward and transition functions, like the estimators of Q-values in ~\citet{osband2016deep}.
In this paper, instead of the model-based approach, we take a model-free one for simplicity. 
We use the RND bonus of state $s'$ as the local uncertainty of $(s,a)$ pair, where $s'$ is the next state after $(s,a)$.
Although the networks in RND can be easily extended to evalute novelty of $(s,a)$ pair instead of $s'$,
we follow the RND original imprementations for a simple and clear comparison.
We discuss the difference between the local uncertainty evaluations in \ref{app:localuncertainty}.
In this case, OPPO is equivalent to RND, if $\beta^2 = c$ and  $c \rightarrow \infty$.
Testing OPPO with various local uncertainty estimators is left for future work.  
We also tested OPPO with local uncertainties based on exact visitation counts of $s'$ i.e., $\frac{1}{n_{s'}}$.

\section{Experiments}\label{sec:experiments}
\subsection{Tabular Domain}
First, we examine the efficiency of the proposed algorithms in a tabular domain where visitation counts are easily calculated.
We used a domain called a bandit tile.
A bandit tile is a kind of a grid world with two tiles exist on which the agent receives a stochastic reward.
We show an example of a bandit tile in \figref{fig:bandit}.
In the figure, `G' represents the tile and `S' represents possible initial positions of the agent.
The initial position is stochastically chosen among the two `S' tiles.
The reward is sampled from a Gaussian distribution.
The mean reward of each `G' tile is $0.5$ and $0.3$ and its variance is $0.5$. 
The episode ends when the agent reaches the `G' tile or 100 time-steps are passed.

We compared OPPO with the bonus based on exact visitation counts to OPPO, RND, and PPO.
\Figref{fig:bandit_results} shows that
OPPO is more efficient than RND and also suggests that we can improve OPPO if there is a proper method to estimate local uncertainty.
\newcommand{\subfigwidth}{5cm}
\newcommand{\width}{5cm}
\begin{figure}[tbh]
\begin{center}
\centerline{\includegraphics[bb=0 0 678 678,width=\width]{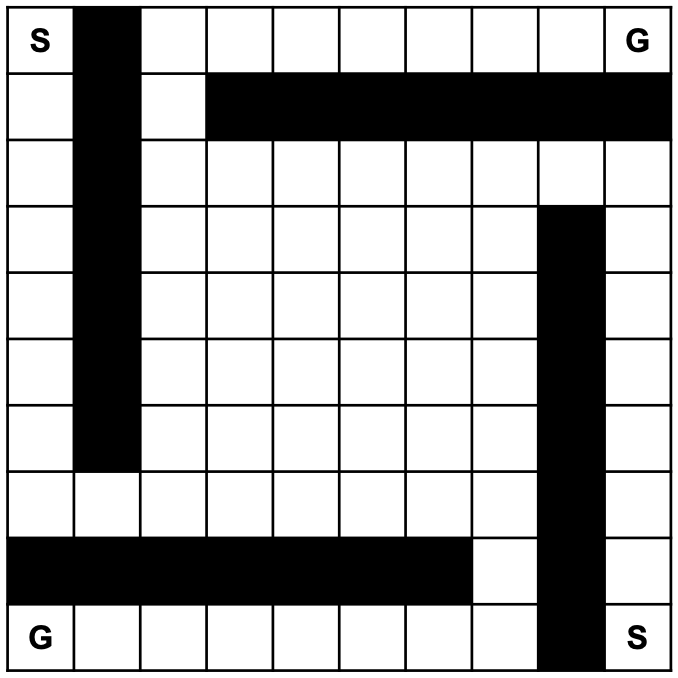}}
\caption{Example of bandit tile domain}
\label{fig:bandit}
\centerline{\includegraphics[bb=0.000000 0.000000 460.800922 345.600691,width=\hsize]{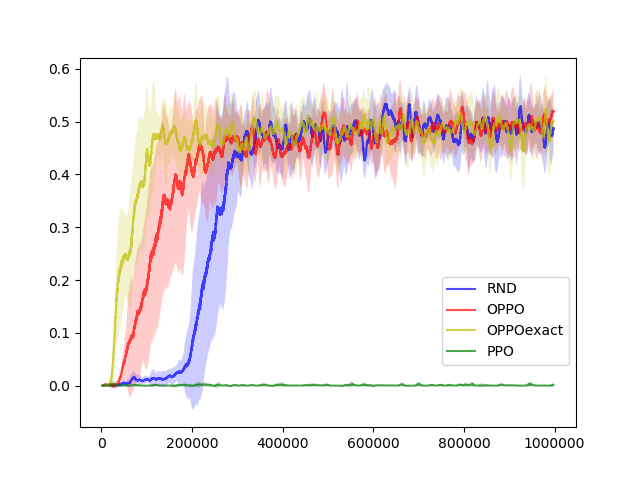}}
\caption{Moving average $\pm$ standard deviation of epsode rewards in bandit tile domain with 10 seeds until 1M time-steps}
\label{fig:bandit_results}
\end{center}
\end{figure}

\subsection{Atari Domain}
Next, we show experimental results on more complex tasks, Atari games, popular testbeds for reinforcement learning.
It has been pointed out that Atari games are deterministic, which is not appropriate for being testbeds, so we added randomness by sticky action~\cite{machado2018revisiting}.
In the sticky action environment, the current chosen action are executed with the probability $1-\zeta$ while the most previous action is repeated with the probability $\zeta$.
We set $\zeta = 1/4$.
We chose six games (Frostbite, Freeway, Solaris, Venture, Montezuma's Revenge, and Private Eye) to evaluate the proposed method and run algorithms until 100 million time-steps in Frostbite and 50 million in the other games. 
OPPO was more effective than RND at Frostbite in terms of learning speed, although the difference is not so salient as that in the tabular case.
The details are shown in \figref{fig:atari} in Appendix.
\comm{ある程度理論の話をしている＋トイプロブレムではポジティブな結果が出ていることを考慮すれば、Atari Domainでさほどポジティブな結果がでてなくても、Workshopは問題ない気がします。
なので、以下のような流れで、手法改良やより洗練された実験は今後していく、というように話を進めてもよいかもしれません（これで本当に良いかはちょっと自信ないですが。。）：
0. Atariのような複雑なドメインでの提案手法の性能を検証するための予備的な実験を行う
1. Frostbiteでややうまくいったことと、ほかではRNDと同程度であること、を主張(したがって、予備的な実験においてはRNDに比べて同等以上の性能を達成)
2. Frostbiteではなぜうまくいったのかを考察
3. RNDと同程度だったのはなぜなのかの考察. (もし手法の改良の余地があればそれを今後の課題として述べる)
}
\comm{
In Atari games, counting state visitation is difficult.
pixel単位でcountしても，ほぼ二度と同じ画面に遷移することは無いので無意味．
したがって，OPPOexactは除いて実験している．

Atari以外で圧勝するタスクがほしい．
Atariは，色々な要素が絡んでいるので，手法の良さを主張するのにやりづらい．
}

\section{Conclusion}
We have proposed a new algorithm, optimisitic proximal policy optimization (OPPO) to alleviate the sparse reward problem.
OPPO is an extension of proximal policy optimization and considers uncertainty of estimation of expected total returns instead of simply estimating the returns.
OPPO optimistically evaluates the values of policies by the amount of uncertainty and improves the policy like PPO.
Experimental results show that OPPO learns more effectively than the existing method, RND, in a tabular domain.
\memo{Frostbiteも入れる？}
%

\comm{
control taskとかで実験してみたい
PopArtみたいなreward normalizationと組み合わせるのは有望かも

workshopの後？  TODO:
bandittileのvarを調節して効きかたを確かめる
control taskも試す

＃Exploration
PixelCNN
を付け加えても良いかも

\subsection{control task}
スパース報酬のコントロールタスクについても実験を行った．
\\

出来合いのスパース報酬問題?

}

\comm{
count-base でやったけど．V値の最終層の出力だと，．．．predictionによる表現の学習は期待できる？
sparseだと，ほぼ常に０なので，大して学習の意味は無いのでは？
}

\clearpage
\section*{Acknowledgments}
Computational resource of AI Bridging Cloud Infrastructure (ABCI) provided by National Institute of Advanced Industrial Science and Technology (AIST) was used.
\bibliographystyle{named}
\bibliography{database}

\clearpage
\appendix
\onecolumn
\section{Details of Proposed Method}
\subsection{Proofs}
\Cororef{coro:relation_var-eta2} is derived from the following relations.
\begin{proof}
\begin{align}
\mathbf{ v a r }_{\tau} \left(\hat{\eta}(\pi) \right)
&= \mathbf{ v a r }_{\tau}\left(\sum_{s,a} \rho(s) \pi(a|s) \hat{Q}^{0,\pi}_{(s,a)} \right)\\
&\leq \sum_{s,a} \rho(s) \pi(a|s) \mathbf{ v a r }_{\tau}\left(\hat{Q}^{0,\pi}_{(s,a)} \right)\\
&\leq \sum_{s,a} \rho(s) \pi(a|s) Q_{2, \tau}^{0, \pi}(s,a))\\
&= \eta_{2,\tau}(\pi)
\end{align}
The first inequality is derived from Jensen's inequality, and the second one is derived from \theoref{theo:uncertaintyBellman}.
\end{proof}

For convenience, we introduce some additional notations.
Let $\rho_{h}^\pi(s)$ denote the probability of the agent being at state $s$ at time-step $h$ under the condition $s_{0} \sim \rho\left(\cdot\right), a_{h} \sim \pi\left(\cdot | s_{h}\right), s_{h+1} \sim T_\tau\left(s_{h}, a_{h}, \cdot\right) \text{for } h \geq 0$ and expectation under the condition as $\mathbb{E}_{s_{0}, a_{0}, \cdots \sim \tilde{\pi}}[\cdot]$.
\theoref{theo:similarity} is derived from the following relations.
\begin{proof}
Firstly, we show that $\eta_{i, \tau}(\pi)$ satisfies the following equations,
\begin{equation}
\eta_{i,\tau} \left( \pi' \right) - \eta_{i,\tau}\left( \pi \right) = \sum _{h,s,a}\rho ^{\pi' }_{h}\left( s\right) \pi' \left( a | s\right) A^{h,\pi }_{i,\tau}\left( s,a\right),
\label{eq:diff_eta}
\end{equation}
which is almost the same as the equations shown in \cite{kakade2002approximately,schulman2015trust}.
\Eqref{eq:diff_eta} is derived as below:
\begin{align}
    \eta_{i, \tau} ( \pi') - \eta_{i, \tau}( \pi )
    &=\E_{s_0, a_0, \dots, \sim \pi'} \left[ \sum ^{H}_{h=0}r\left( s_h,a_{h}\right) - V^{0,\pi}_{i,\tau} \left( s_{0}\right) \right]\\
&=\E_{s_0, a_0, \dots, \sim \pi'} \left[ \sum ^{H}_{h=0}  \left\{r( s_h,a_{h}) +V^{h+1,\pi}_{i,\tau} ( s_{h+1}) - V^{h,\pi}_{i,\tau} ( s_{h}) \right\} \right]\\
&=\E_{s_0, a_0, \dots, \sim \pi'} \left[\sum ^{H}_{h=0} A^{h,\pi}_{i,\tau} \left( s_{h}, a_h\right) \right]\\
&=\sum_{h, s, a} \rho_h^{\pi'}(s) \pi'(a|s)A^{h,\pi}_{i,\tau} \left( s, a\right).
\end{align}
The first equation is derived from the definition of $\eta$ and the fact that sampling of the initial state only depends on $\rho(\cdot)$,
the second one $V^{H+1} = 0$.
The third one and the forth one are derived from the definition of $A^{h,\pi}_{i, \tau}$ and $\E_{s_0, a_0, \dots, \sim \pi'}[\cdot]$, respectively.

For simplicity, we denote $\pi_\phi$ as $\pi$.
By the fact that $\sum_{a} \pi(a|s)  {A}_{i, \tau}^{h,\pi}(s,a) = 0 \; (i = 1,2)$, 
\begin{align}
  \mathcal{L}_{\tau}(\pi, \pi) - \tilde{\eta}_{\tau}(\pi)  &=   \sum_{h,s, a}\rho_h^{\pi}(s) \pi(a|s) \left(  {A}_{1, \tau}^{h,\pi}(s,a) + \beta \frac{{A}_{2, \tau}^{h,\pi}(s,a)}{\sqrt{\eta_{2,\tau}(\pi)}} \right)\\
  &= 0.
\end{align}
Also, 
\begin{align}
  \nabla_\theta \mathcal{L}_{\tau}(\pi_\phi, \pi_\theta)|_{\theta = \phi} - \nabla_\theta \tilde{\eta}_{\tau}(\pi_\theta) |_{\theta = \phi}
  &= \left.\nabla_\theta \sum_{h,s, a}\rho_h^{\pi}(s) \pi_\theta(a|s) \left(  {A}_{1, \tau}^{h,\pi}(s,a) + \beta \frac{{A}_{2, \tau}^{h,\pi}(s,a)}{\sqrt{\eta_{2,\tau}(\pi)}} \right) \right|_{\theta = \phi} \nonumber \\
  &- \left.\nabla_\theta \sum_{h,s, a}\rho_h^{\pi_\theta}(s) \pi_\theta(a|s) \left(  {A}_{1, \tau}^{h,\pi}(s,a) + \beta \frac{{A}_{2, \tau}^{h,\pi}(s,a)}{\sqrt{\eta_{2,\tau}(\pi)}} \right) \right|_{\theta = \phi}\\
  &= -\sum_{h,s}\nabla_\theta \rho_h^{\pi_\theta}(s) \rvert_{\theta = \phi} \sum_a \pi(a|s) \left(  {A}_{1, \tau}^{h,\pi}(s,a) + \beta \frac{{A}_{2, \tau}^{h,\pi}(s,a)}{\sqrt{\eta_{2,\tau}(\pi)}} \right)\\
  &= 0.
\end{align}
The first equation is derived from equation~\eqref{eq:diff_eta}.
\end{proof}

\subsection{Algorithm}\label{app:alg}
\memo{$\tau$を省くのか省かないのか}
In the batch data, we denote the state, action, and reward at time-step $h \; (0\leq h \leq T)$ and sampled by actor $n \; (0 \leq n \leq N-1)$ are $s_h^{(n)}, a_h^{(n)}$, and $r_h^{(n)}$, respectively.
Let $r_{1,h}^{(n)}$ denote $r_{h}^{(n)}$ and $r_{2,h}^{(n)}$ denote the local uncertainty of $(s_h^{(n)}, a_h^{(n)})$.
$A_i \; (i = 1, 2)$ in equation~\eqref{def:oppoobj} is calculated as below:
\begin{align}
A_i (s_l^{(n)}, a_l^{(n)}) = \sum_{h=l}^{T-1} (\gamma^i \lambda)^{h-l} \left\{\gamma^i V_i(s_{h+1}^{(n)}) + r_{i,h}^{(n)} - V_i(s_{h}^{(n)})\right\},
\end{align}
where $V_i$ is an estimator of $V_{i, \tau}^{\pi}$ and  $\gamma$ is a discount factor.
The discount factor is often used even if the horizon is finite, so we follow the ordinary implementations. 
$\eta_2$ is calculated as below:
\begin{align}
\eta_2 = \sum_{n = 0}^{N-1} V_2(s_{0}^{(n)}) + A_2(s_{0}^{(n)}, a_{0}^{(n)})
\end{align}

Pseudo code is shown at Algorithm~\ref{alg:algorithm}.
\memo{書き足す}

\subsection{Local Uncertainty Estimation}\label{app:localuncertainty}
Let $\nu(s')$ denote the local uncertainty based on the next state $s'$ after $(s,a)$ pair.
OPPO uses $\nu(s')$ as the local uncertainty of $(s,a)$ instead of $\nu(s,a)$.
There is a small gap between the discussion and the implementation of OPPO.
However, using $\nu(s')$ is reasonable if the state transition is a tree, a graph without cycles.
Using $\nu(s')$ means using the average of $\nu(s')$ as the local uncertainty of $(s,a)$.
This can be approximated by $\sum_{s'} T(s,a,s')\nu(s')$.
In the tree case, $n_{s'}$ can be approximated by $T(s,a,s')n_{s,a}$.
Thus, if $\nu(s') \approx \frac{1}{n_{s'}}$,
the local uncertainty of $(s,a)$ can be approximated by $\sum_{s'} T(s,a,s')\nu(s') \approx \sum_{s'} T(s,a,s')\frac{1}{n_{s'}} \approx \sum_{s'} \frac{1}{n_{s,a}} = \frac{|\mathcal{S}_{s,a}|}{n_{s,a}}$.
This means that $\nu(s,a)$ can be approximated by the average of $\nu(s')$, if $C_u = |\mathcal{S}_{s,a}|$.

\begin{algorithm}[tb]
\caption{OPPO}
\label{alg:algorithm}
\begin{algorithmic}[1] 
\STATE initialize the parameters of the policy network, the V-value estimators and the local uncertainty estimator.
\FOR{ $\tau = 0, \dots $}
\FOR{ $n = 0, \dots N-1$}
\FOR{ $t = 0, \dots T-1$}
\STATE make a batch data $\tau$ by sampling action $a_t^{(n)}$ from $\pi(\cdot|s_t^{(n)})$, exectuting $a_t^{(n)}$, and receiving next state $s_{t+1}^{(n)}$, extrinsic reward $r_t^{(n)}$, and the local uncertainty of $(s_t^{(n)},a_t^{(n)})$.
\ENDFOR
\ENDFOR
\STATE update policy so as to maximize the objective function of OPPO based on the data. 
\ENDFOR
\end{algorithmic}
\end{algorithm}
\memo{擬似コードは雑すぎる？ツッコミが来たら対応すれば良いか．．．}
\memo{面倒なので，$r_\tau$と$\nu_\tau$の代わりに$r_{1,\tau}$と$r_{2,\tau}$とかという定義にしておく？}

\section{Further Investigation in Tabular Domain}
\begin{figure}[ht]
\vskip 0.2in
\begin{center}
\centerline{\includegraphics[bb=0.000000 0.000000 460.800922 345.600691,width=0.5\hsize]{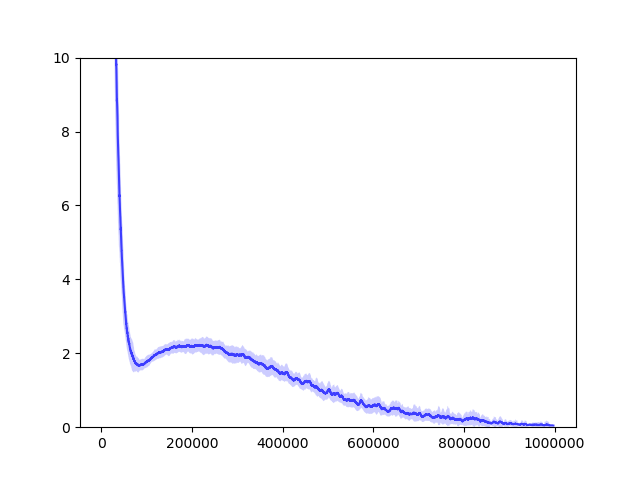}}
\caption{Moving average of the average of $\frac{\text{RND bonus}}{ 1/n_{s'}}$ in batch data.}
\label{fig:rateint}
\end{center}
\vskip -0.2in
\end{figure}

To confirm the validity of using RND bonus as visitation counts, we measured a ratio $\frac{\text{RND bonus}}{ 1/n_{s'}}$ to check if it is stable at around one in the bandit tile domain. 
\Figref{fig:rateint} shows that the ratio was around 1 for millions of time-steps, although it was high at the beginning and nearly zero at the end.
It can be considered that OPPO is worse than OPPO with the exact count bonus by the amount of the overvaluation, and that the undervaluation was not harmful because it occured after learning the policy to the best tile.

\section{Details of Results in Atari Games}
We compared OPPO with RND in the six Atari games.
In the original RND implementation, a reward clipping technique which transforms negative/positive extrinsic reward to $\{-1, 1\}$ is used, so we also used this technique in OPPO and RND.
Note that we use a frame skipping technique, and the number of the frame skips is four; so one time-step is equal to or less than four frames (it is less than four if the episode ends at a skipped frame).

\begin{figure*}[th]
  \begin{subfigmatrix}{2}
 \subfigure[Frostbite]{
   \includegraphics[bb=0.000000 0.000000 460.800922 345.600691, width=0.48\hsize]{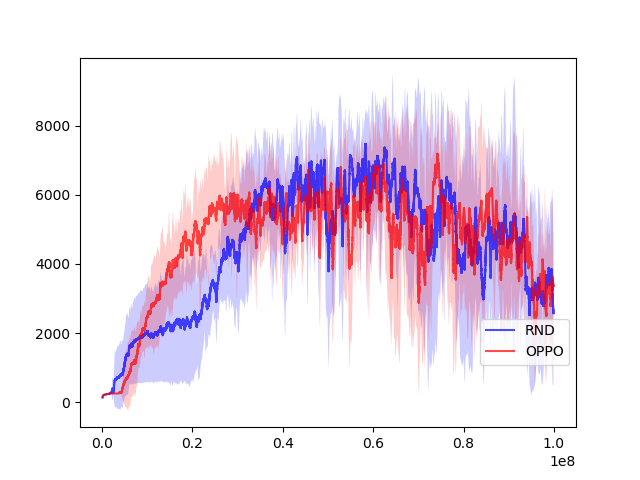}}
 \subfigure[Freeway]{
   \includegraphics[bb=0.000000 0.000000 460.800922 345.600691, width=0.48\hsize]{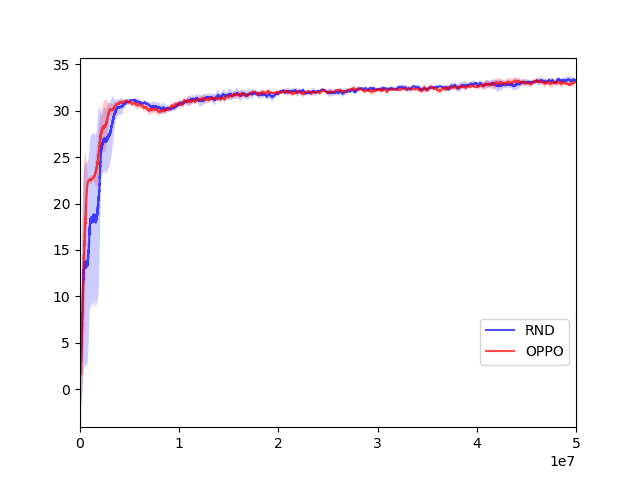}}
 \subfigure[Solaris]{
   \includegraphics[bb=0.000000 0.000000 460.800922 345.600691, width=0.48\hsize]{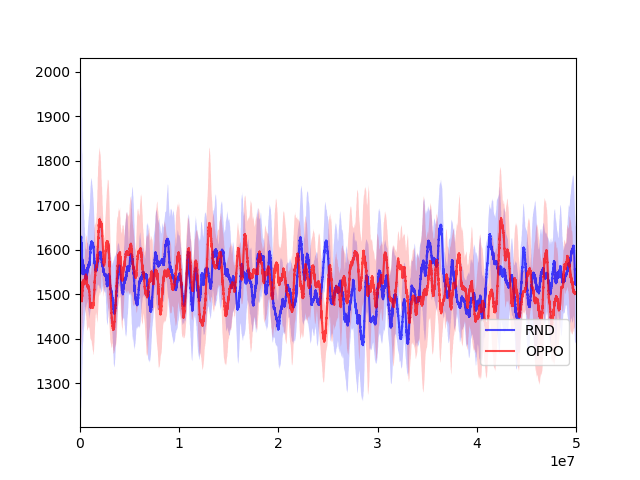}}
 \subfigure[Venture]{
   \includegraphics[bb=0.000000 0.000000 460.800922 345.600691, width=0.48\hsize]{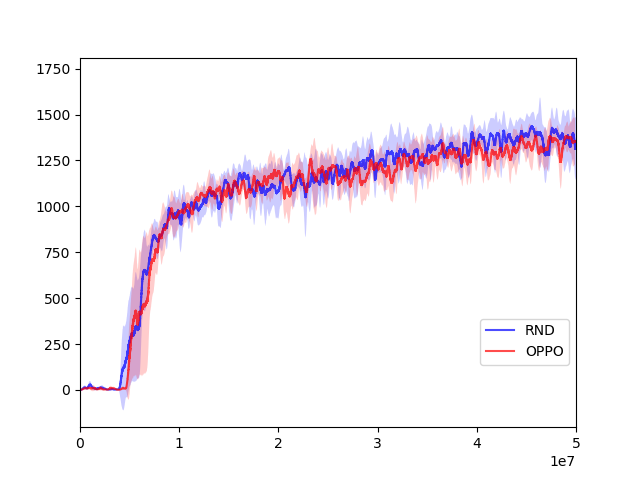}}
 \subfigure[Montezuma's Revenge]{
   \includegraphics[bb=0.000000 0.000000 460.800922 345.600691, width=0.48\hsize]{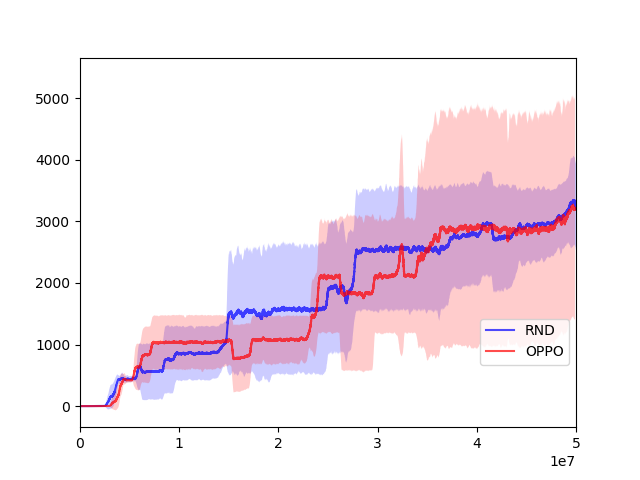}}
 \subfigure[Private Eye]{
   \includegraphics[bb=0.000000 0.000000 460.800922 345.600691, width=0.48\hsize]{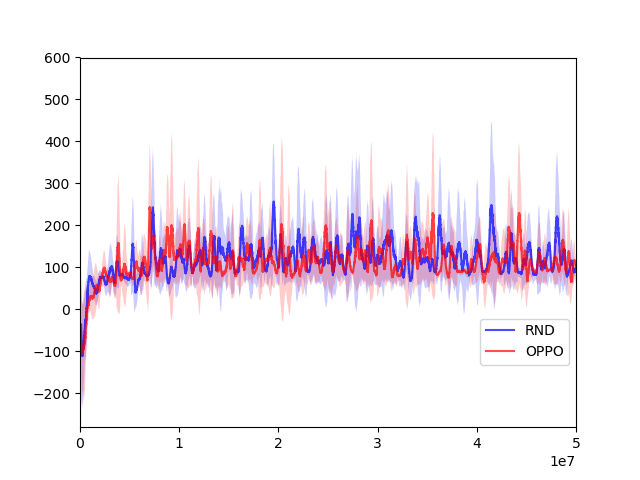}}
 \end{subfigmatrix}
\caption[]{Moving average $\pm$ standard deviation of episode rewards with 5 seeds until 50M time-steps (100M time-steps in Frostbite)}
  \label{fig:atari}
\end{figure*}

\Figref{fig:atari} shows that OPPO learns more effectively than RND in Frostbite although there is only slight difference with the other games.
Also, \Figref{fig:atari} shows that exrinsic rewards decrease in Frostbite.
One of the reason for the decrease may be the reward clipping, although further investigation is needed to confirm that.
By the reward clipping, the agent learns a policy to receive positive rewards with high frequency, not high returns.
The agent may learn the policy with the same frequency of rewards but with a small total return, as it receives data.
Note that there are small and large rewards in Frostbite, and that a novel states leads to a higher reward in most Atari games~\cite{burda2019large}. 
This problem can be alleviated by rescaling the reward by considering the amount of reward, e.g. PopArt~\cite{van2016learning}, which is left for future work.

\comm{
reward clipありの設定でも，正の報酬が得られる頻度の違いを見極めるのが本質的に重要なドメインなら，差がつくはず．
複数の行動に対して，報酬の得られる頻度の期待値が異なり，それが本質的な差を生むようなドメインでは効果的？
Frostbiteでは，流氷から落ちやすい/にくい位置への移動とか，Freewayでは車に轢かれやすい/にくい位置への移動とかが本質的に重要？
その一方でMontezuma等は，randomnessが本質的には効いてこない？（sticky actionによって経路にゆらぎがあっても，．．．ドクロに当たるとか割と本質的に効くのでは？でも報酬に直接効く訳でもない．．．）
}


\end{document}